\documentclass[runningheads]{llncs}
\usepackage[T1]{fontenc}
\usepackage{graphicx}
\usepackage{hyperref}
\usepackage{color}
\usepackage{amsmath}
\usepackage{booktabs}
\usepackage{multirow}
\usepackage{enumitem}

\usepackage{tablefootnote}
\usepackage{adjustbox} 
\usepackage{tabularx}
\newcolumntype{Y}{>{\centering\arraybackslash}X}
\usepackage{bbold} 
\usepackage{arydshln} 
\usepackage{caption} 
\usepackage{cleveref} 
\usepackage{makecell} 

\begin{document}

\title{Large Scale Supervised Pretraining For Traumatic Brain Injury Segmentation}
\titlerunning{Large Scale Supervised Pretraining For TBI Segmentation}
%


\author{Constantin Ulrich\inst{1,4,5} \and
Tassilo Wald\inst{1,2,6} \and
Fabian Isensee\inst{1,2} \and
Klaus H. Maier-Hein\inst{1,3}}

\authorrunning{C. Ulrich et al.}

\institute{German Cancer Research Center (DKFZ), Heidelberg, Division of Medical Image Computing, Germany\\
\and Helmholtz Imaging, DKFZ, Heidelberg, Germany\\
\and Pattern Analysis and Learning Group, Department of Radiation Oncology, Heidelberg University Hospital, Heidelberg, Germany \\
\and National Center for Tumor Diseases (NCT), NCT Heidelberg, A partnership between DKFZ and University Medical Center Heidelberg\\
\and Medical Faculty Heidelberg, University of Heidelberg, Heidelberg, Germany \\
\and Faculty of Mathematics and Computer Science, Heidelberg University, Germany \\
\email{constantin.ulrich@dkfz-heidelberg.de}}
\maketitle 
\begin{abstract}
The segmentation of lesions in Moderate to Severe Traumatic Brain Injury (msTBI) presents a significant challenge in neuroimaging due to the diverse characteristics of these lesions, which vary in size, shape, and distribution across brain regions and tissue types. This heterogeneity complicates traditional image processing techniques, resulting in critical errors in tasks such as image registration and brain parcellation. To address these challenges, the AIMS-TBI Segmentation Challenge 2024 aims to advance innovative segmentation algorithms specifically designed for T1-weighted MRI data, the most widely utilized imaging modality in clinical practice.

Our proposed solution leverages a large-scale multi-dataset supervised pretraining approach inspired by the MultiTalent method. We train a Resenc L network on a comprehensive collection of datasets covering various anatomical and pathological structures, which equips the model with a robust understanding of brain anatomy and pathology. Following this, the model is fine-tuned on msTBI-specific data to optimize its performance for the unique characteristics of T1-weighted MRI scans and outperforms the baseline without pretraining up to 2 Dice points.

\end{abstract}

\section{Introduction}

The segmentation of lesions in Moderate to Severe Traumatic Brain Injury (msTBI) represents a challenging task in neuroimaging, making it an intriguing and compelling competition for the research community. Unlike more common brain pathologies such as stroke or tumors, msTBI lesions are exceptionally diverse, varying significantly in size, shape, and distribution across different brain regions and tissue types. This heterogeneity is a hallmark of msTBI, and it complicates conventional image processing techniques, often leading to substantial errors in critical tasks like image registration and brain parcellation \cite{Diamond2020-kl,King2020-zf}. 

The complexity of msTBI lesions, combined with their impact on multiple aspects of brain function and structure, makes this segmentation task particularly difficult and, therefore, an ideal candidate for a competitive challenge for the medical image processing community. Existing segmentation tools, designed for other types of brain injuries, struggle with the specific characteristics of msTBI, especially when only T1-weighted MRI scans are available \cite{HENSCHEL2020117012,Jain2019-im}. This limitation is significant because T1-weighted MRI is a very popular modality across a large field of clinical applications and the most used modality in the ENIGMA consortium\footnote[1]{www.enigma.ini.usc.edu}. 

The AIMS-TBI segmentation challenge 2024 is designed to push the development novel segmentation algorithms that can effectively handle the intricate and varied nature of msTBI lesions. By focusing exclusively on T1-weighted MRI data, this competition underscores the need for solutions that can be broadly applied across multiple clinical sites. 
Advances in lesion segmentation, enabling the use of highly accurate lesion masks in downstream analyses such as parcellation, functional connectivity, connectomics, and fixel-based studies, will improve prognostic accuracy and have the potential to greatly enhance long-term patient outcomes. This makes the msTBI lesion segmentation challenge not only a test of technical skill but also a meaningful contribution to the future of neuroimaging. 

For our challenge solution, we propose to leverage a large-scale multi-dataset supervised pretraining inspired by MultiTalent \cite{multitalent}, which has been successful in other complex segmentation tasks. Our solution involves pretraining a deep learning model on a vast, diverse collection of datasets, encompassing various image modalities and a variety of pathological and anatomical structures. This extensive pretraining allows the model to develop a robust understanding of the human body, the brain anatomy, and corresponding pathologies, enabling it to generalize effectively to the diverse lesion types seen in the challenge dataset. 
The pretrained model is then fine-tuned on msTBI-specific data, optimizing it for the nuances of TBI lesions in T1-weighted MRI scans. 
We evaluate the performance of our proposed solution across different age groups, time since injury, and gender.

\section{Methods}

\subsection{Challenge Dataset}
The challenge dataset consists of almost 763 T1-weighted MRI scans, divided into 388 training, 100 validation, and 275 test cases \cite{Dennis2024}. Data were collected from 12 different cohorts, ensuring a balanced distribution of cases across training, validation, and test sets. The MRI scans were acquired using both 1.5T and 3T scanners. The segmentation process involved dual annotation by 7 primary raters from the University of Utah, trained under a standardized protocol, and reviewed by 5 expert raters. Before working on real data, the primary raters were required to achieve a Dice Similarity Coefficient (DSC) of 0.6. Lesions were identified in a three-step process involving manual segmentation, review by a second rater, and final approval by an expert. Pre-processing of the dataset involved defacing the MRI images to protect subject privacy.

\subsection{Large Supervised Pretraining}
We employed a large-scale supervised pretraining for this challenge that draws inspiration from the MultiTalent approach, utilizing a separate segmentation head for each pretraining dataset \cite{multitalent}. We gathered a comprehensive and diverse collection of publicly available segmentation datasets, covering various anatomical and pathological structures, and including multiple imaging modalities such as Computer Tomography (CT), Positron Emission tomography (PET), and various Magnetic Resonance Imaging (MRI) sequences (detailed in Table \ref{tab:datasets}). 
The entire pretraining and fine-tuning pipeline was implemented using the nnU-Net framework \cite{isensee_nnu-net_2021}, with the architecture based on the recently introduced ResencL U-Net \cite{extendingnnunet,revisited}.
The model was trained for 4000 epochs with a patch size of [192,192,192] and a batch size of 24. All images were resampled to a cubic 1mm resolution and Z-score normalized. Consistent with the MultiTalent method, datasets were sampled inversely proportional to the square root of the number of images per dataset to ensure balanced training. The resulting foundation model can effectively segment all datasets simultaneously and demonstrates strong potential for fine-tuning on specific downstream tasks. It should be noted that this model was not explicitly trained for this challenge. The public available model weights can be found \hyperlink{here}{https://zenodo.org/records/13753413}.

\subsection{Finetuning}
Given the large pretrained model, we fine-tuned it directly on the challenge dataset. We applied the same preprocessing as in pretraining, including resampling to 1mm isotropic resolution and Z-score normalization. The segmentation heads were randomly initialized, while all other parameters were inherited from the pretrained network. We adhered to the default nnU-Net training scheme, utilizing a combination of Cross Entropy and Dice loss, and a poly learning rate schedule with an initial learning rate of 0.01 \cite{isensee_nnu-net_2021}. Additionally, we experimented with the following fine-tuning schedules:

\begin{itemize}[label=\textbullet]
    \item Default settings with an initial learning rate of 0.001
    \item A linear learning rate increase over 50 epochs, up to 0.01, followed by the default schedule
    \item A linear learning rate increase over 50 epochs, up to 0.001, followed by the default schedule
\end{itemize}

\begin{table}[t]
\centering
\caption{Overview over all datasets used for the supervised pretraining}
\label{tab:datasets}
\resizebox{\textwidth}{!}{ 
\begin{tabular}{lccrr}
\toprule
Name & Images & Modality & Target & Link \\
\midrule
Decatlon Task 2 \cite{antonelli2021medical,simpson2019large}& 20 & MRI & Heart & \url{http://medicaldecathlon.com/} \\
Decatlon Task 3 \cite{antonelli2021medical,simpson2019large}& 131 & CT & Liver, L. Tumor & \url{http://medicaldecathlon.com/} \\
Decatlon Task 4 \cite{antonelli2021medical,simpson2019large}& 208 & MRI & Hippocampus & \url{http://medicaldecathlon.com/} \\
Decatlon Task 5 \cite{antonelli2021medical,simpson2019large}& 32 & MRI & Prostate & \url{http://medicaldecathlon.com/} \\
Decatlon Task 6 \cite{antonelli2021medical,simpson2019large}& 63 & CT & Lung Lesion & \url{http://medicaldecathlon.com/} \\
Decatlon Task 7 \cite{antonelli2021medical,simpson2019large}& 281 & CT & Pancreas, P. Tumor & \url{http://medicaldecathlon.com/} \\
Decatlon Task 8 \cite{antonelli2021medical,simpson2019large}& 303 & CT & Hepatic Vessel, H. Tumor & \url{http://medicaldecathlon.com/} \\
Decatlon Task 9 \cite{antonelli2021medical,simpson2019large}& 41 & CT & Spleen & \url{http://medicaldecathlon.com/} \\
Decatlon Task 10 \cite{antonelli2021medical,simpson2019large}& 126 & CT & Colon Tumor & \url{http://medicaldecathlon.com/} \\
ISLES2015 \cite{isles2015}& 28 & MRI & Stroke Lesion & \url{http://www.isles-challenge.org/ISLES2015/} \\
BTCV \cite{BTCV}& 30 & CT & 13 abdominal organs & \url{https://www.synapse.org/Synapse:syn3193805/wiki/89480} \\
LIDC \cite{lidc}& 1010 & CT & Lung lesion & \url{https://www.cancerimagingarchive.net/collection/lidc-idri/} \\
Promise12 \cite{Litjens2014}& 50 & MRI & Prostate & \url{https://zenodo.org/records/8026660} \\
ACDC \cite{Bernard2018}& 200 & MRI & RV cavity, myocardium, LV cavity & \url{https://www.creatis.insa-lyon.fr/Challenge/acdc/databases.html} \\
ISBILesion2015 \cite{Carass2017}& 42 & MRI & MS Lesion & \url{https://iacl.ece.jhu.edu/index.php/MSChallenge} \\
CHAOS \cite{CHAOSdata2019}& 60 & MRI & Liver, Kidney (L\&R), Spleen & \url{https://zenodo.org/records/3431873} \\
BTCV 2 \cite{BTCV2}& 63 & CT & 9 abdominal organs & \url{https://zenodo.org/records/1169361\#.YiDLFnXMJFE} \\
StructSeg Task1 \cite{structseg} & 50 & CT & 22 OAR Head \& neck & \url{https://structseg2019.grand-challenge.org} \\
StructSeg Task2 \cite{structseg} & 50 & CT & Nasopharynx cancer & \url{https://structseg2019.grand-challenge.org/Home/} \\
StructSeg Task3 \cite{structseg} & 50 & CT & 6 OAR Lung & \url{https://structseg2019.grand-challenge.org/Home/} \\
StructSeg Task4 \cite{structseg} & 50 & CT & Lung Cancer & \url{https://structseg2019.grand-challenge.org/Home/} \\
SegTHOR \cite{lambert2019segthor} & 40 & CT & heart, aorta, trachea, esophagus & \url{https://competitions.codalab.org/competitions/21145} \\
NIH-Pan \cite{roth2015deeporgan,clark_cancer_2013,NHpancreas}  & 82 & CT & Pancreas & \url{https://wiki.cancerimagingarchive.net/display/Public/Pancreas-CT} \\
VerSe2020 \cite{verse1,verse2,verse3} & 113 & CT & 28 Vertebrae & \url{https://github.com/anjany/verse} \\
M\&Ms \cite{Campello2021,10103611}& 300 & MRI & l. ventricle, r. ventricle, l. ventri. myocardium & \url{https://www.ub.edu/mnms/} \\
ProstateX \cite{prostatex} & 140 & MRI & Prostate lesion & \url{https://www.aapm.org/GrandChallenge/PROSTATEx-2/} \\
RibSeg \cite{ripseg}& 370 & CT & Rips & \url{https://github.com/M3DV/RibSeg?tab=readme-ov-file} \\
MSLesion \cite{Muslim2022-qt} & 48 & MRI & MS Lesion & \url{https://data.mendeley.com/datasets/8bctsm8jz7/1} \\
BrainMetShare \cite{Grovik2020-un} & 84 & MRI & Brain Metastases & \url{https://aimi.stanford.edu/brainmetshare} \\
CrossModa22 \cite{Shapey2021-iz}& 168 & MRI & vestibular schwannoma, cochlea & \url{https://crossmoda2022.grand-challenge.org/} \\
Atlas22 \cite{Liew2018-wy}& 524 & MRI & stroke lesion & \url{https://atlas.grand-challenge.org/} \\
KiTs23 \cite{heller2023kits21} & 489 & CT & Kidneys, k. Tumors, Cysts & \url{https://kits-challenge.org/kits23/} \\
AutoPet2 \cite{Gatidis2022-ms}& 1014 & PET,CT & Lesions & \url{https://autopet-ii.grand-challenge.org/} \\
AMOS \cite{ji2022amos}& 360 & CT,MRI & 15 abdominal organs & \url{https://amos22.grand-challenge.org/} \\
BraTs23 \cite{Karargyris2023,bratsgli1,bratsgli2,bratsgli3} & 1251 & MRI & Glioblastoma & \url{https://www.synapse.org/Synapse:syn51156910/wiki/621282} \\
AbdomenAtlas1.0 \cite{li2024well,qu2023abdomenatlas}& 5195 & CT & 8 abdominal organs & \url{https://github.com/MrGiovanni/AbdomenAtlas?tab=readme-ov-file} \\
TotalSegmentatorV2 \cite{totalseg}& 1180 & CT & 117 classes of whole body & \url{https://github.com/wasserth/TotalSegmentator} \\
Hecktor2022 \cite{Andrearczyk2023-ii}& 524 & PET,CT & nodal Gross Tumor Volumes (Head\&Neck) & \url{https://hecktor.grand-challenge.org/} \\
FLARE \cite{FLARE22} & 50 & CT & 13 abdominal organs & \url{https://flare22.grand-challenge.org/} \\
SegRap \cite{Luo2023-zk}& 120 & CT & 45 OARs (Head\&Neck) & \url{https://segrap2023.grand-challenge.org/} \\
SegA \cite{Radl2022-fz,Jin2021-jo,Pepe2020-fh}& 56 & CT & Aorta & \url{https://multicenteraorta.grand-challenge.org/data/} \\
WORD \cite{luo2022word,liao2023comprehensive}& 120 & CT & 16 abdominal organs & \url{https://github.com/HiLab-git/WORD} \\
AbdomenCT1K \cite{Ma-2021-AbdomenCT-1K} & 996 & CT & Liver, Kidney, Spleen, pancreas & \url{https://github.com/JunMa11/AbdomenCT-1K} \\
DAP-ATLAS \cite{jaus2023towards}& 533 & CT & 142 classes of whole body & \url{https://github.com/alexanderjaus/AtlasDataset} \\
CTORG \cite{Rister2019-tm} & 140 & CT & lung, brain, bones, liver, kidneys and bladder & \url{https://www.cancerimagingarchive.net/collection/ct-org/} \\
HanSeg \cite{Podobnik2023-mp}& 42 & CT & OAR (Head\&Neck) & \url{https://han-seg2023.grand-challenge.org/} \\
TopCow \cite{topcowchallenge}& 200 & CT,MRI & vessel components of CoW & \url{https://topcow23.grand-challenge.org/} \\
\bottomrule
\end{tabular}
}
\end{table}

\section{Results}

\subsection{Development Results}
Table \ref{tab:5foldtable} presents the results of the 5-fold cross-validation conducted on the training dataset generated during model development. The results demonstrate a consistent improvement in performance when pretraining is utilized. The best pretrained model achieves an average Dice score of 54.21, compared to a baseline without pretraining of 52.54. In general, it can be observed that fine-tuning with a smaller learning rate yields better results, as it likely prevents the model from forgetting previously learned feature representations. We selected the two best-performing methods for the final challenge submission and submitted an ensemble of the two methods. However, due to time constraints imposed by the challenge organizers, we could not apply the default test-time augmentations during inference. 

\begin{table}[!htp]\centering
\caption{Dice results of a 5-fold cross-validation using different fine-tuning strategies.}\label{tab:5foldtable}
\scriptsize
\begin{tabular}{llcccccc}\toprule
Architecture &FT schedule &fold 0 &fold 1 &fold 2 &fold 3 &fold 4 &Average \\\midrule
3D U-Net & no pretraining &54.22 &53.54 &55.18 &43.09 &\textbf{56.66} &52.54 \\
3D ResencL U-Net & no pretraining &54.62 & \textbf{58.08} & 54.95& 43.42& 56.15& 53.44 \\
3D ResencL U-Net & lr 0.001 & \underline{57.65} & \underline{55.80} & 55.39 & 43.59 & \underline{56.57} & \underline{53.80} \\
3D ResencL U-Net & warm-up + lr 0.01 & 56.86 & 51.53 & \underline{55.62} & \textbf{47.35} & 55.03 & 53.28 \\
3D ResencL U-Net & warm-up + lr 0.001 & \textbf{58.62} & 54.30 & \textbf{56.36} & \underline{45.40} & 56.36 & \textbf{54.21} \\
\bottomrule
\end{tabular}
\end{table}

\subsection{Test Results}
\begin{figure}
    \centering
    \includegraphics[width=0.99\textwidth]{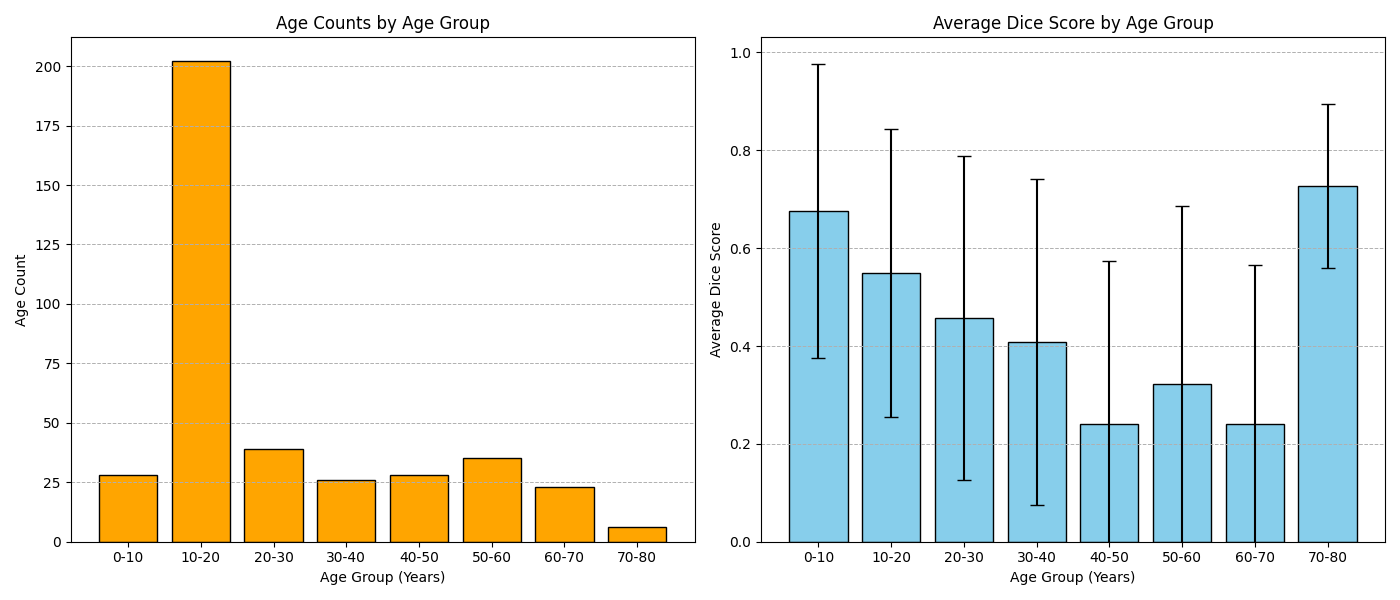}
    \caption{Prediction performance by age group (right) and the number of patients per age group in the dataset (left). We excluded all cases where no ground truth foreground was annotated, and our method made no predictions.}
    \label{fig:age}
\end{figure}

\begin{figure}
    \centering
    \includegraphics[width=0.99\textwidth]{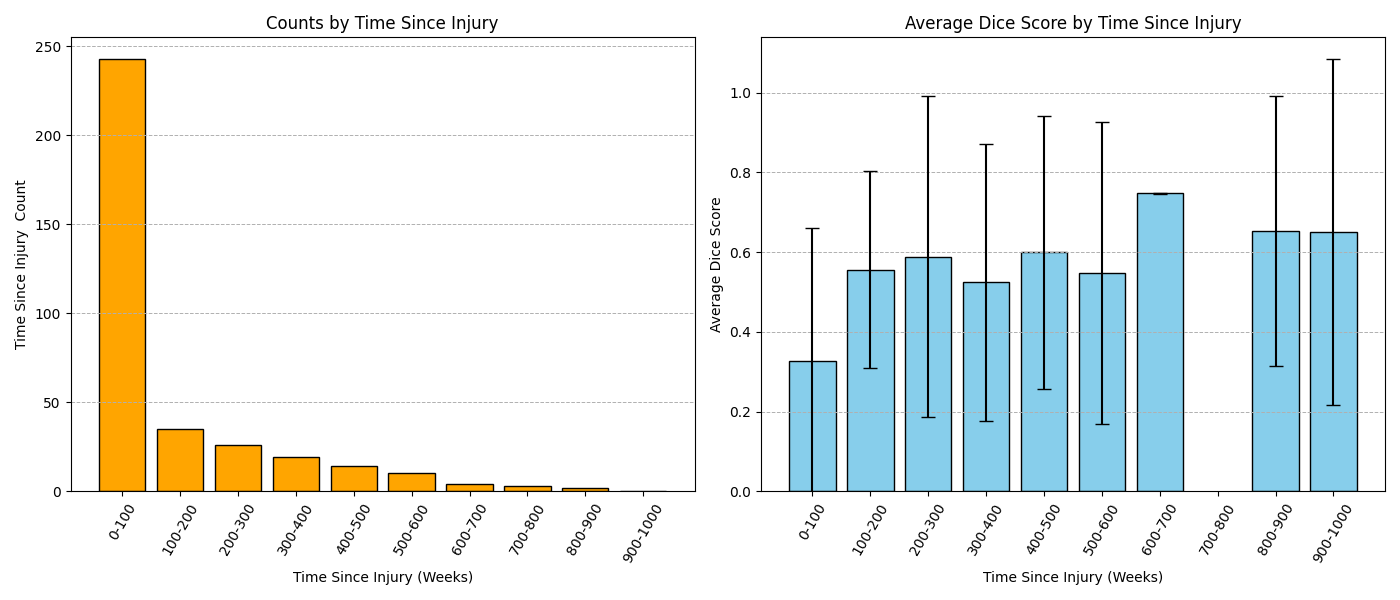}
    \caption{Prediction performance by Time Since Injury (TSI) group (right) and the number of patients per TSI group in the dataset (left). We excluded all cases where no ground truth foreground was annotated, and our method made no predictions.}
    \label{fig:tsi}
\end{figure}

\begin{figure}
    \centering
\includegraphics[width=0.99\textwidth]{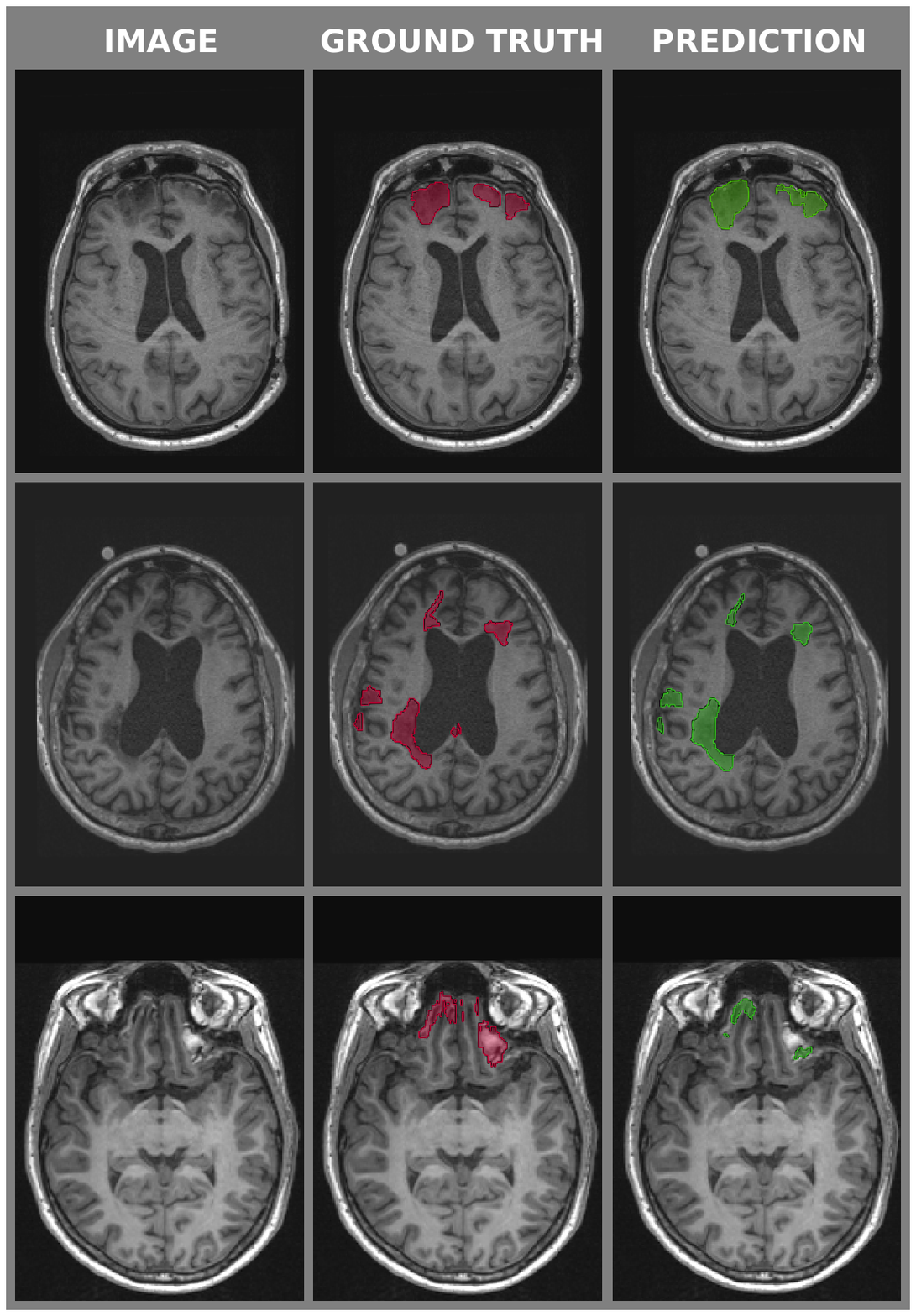}
\caption{Qualitative Segmentation Results: The segmentation results for the first two patients closely align with the ground truth annotations. In contrast, the last image was chosen due to a significant discrepancy between the prediction and the ground truth.}
    \label{fig:quality}
\end{figure}

After the final submission, the organizers provided the test dataset, which included demographic information. In this section, we analyze the performance of our submitted ensemble model. Figure \ref{fig:quality} presents several qualitative predictions alongside the corresponding ground truth annotations. 
The challenge employed an aggregation method where cases without ground truth were assigned a Dice score of 100 if nothing was predicted, but a score of 0 if any prediction was made. In contrast, nnU-Net's evaluation method excludes cases with no foreground and no prediction. This difference in aggregation strategies can significantly affect the final metric, as demonstrated in Table \ref{tab:testresults}. \\
The results differ from our final challenge solution due to the use of float16 precision when ensembling the predicted probabilities from the two models. Unfortunately, this introduced numerical errors that negatively impacted the results. The corrected version would have outperformed the best solution of the test set by 1 dice point. \\
The test dataset comprised 175 male and 100 female cases, while the training set included 231 male and 157 female cases. Unsurprisingly, the method performed better on male patients due to this imbalance. This highlights the importance of using a balanced dataset for clinical applications, as gender disparity can impact model performance. 
"Figure \ref{fig:age} illustrates the Dice score in relation to patient age. The majority of patients were aged between 10 and 20 years. A decline in performance is noticeable in older age groups, with the exception of the last bar, representing patients aged 70-80, which is less informative due to the small sample size. Due to the high variation in the results, our observations regarding age-related performance differences are not statistically significant."
Figure \ref{fig:tsi} presents the Dice score in relation to Time Since Injury (TSI). In this case, no clear trend is observed based on the number of patients across different times since injury.

\begin{table}[]
\centering
\caption{Dice and Normalized Surface Dice (NSD, 1mm tolerance) test set results: proposed model ensembling (left) vs. submitted results with numerical error (right)}\label{tab:testresults}
\scriptsize
\begin{tabular}{lcccc}\toprule
Aggregation & Group & Dice & NSD \\ \midrule
  NaN = 1 & Male & 64.02 & 73.11 \\
NaN = 1 & Female & 58.85 & 66.16 \\
 NaN = 1 & All &  62.14 & 70.59\\
  ignore Nan & Male & 41.71 & 56.43 \\
 ignore Nan & Female & 38.58 & 49.50 \\
 ignore Nan & All & 40.51 & 53.78 \\

\bottomrule
\end{tabular}
\quad
\begin{tabular}{lcccc}\toprule
Aggregation & Group & Dice & NSD \\ \midrule
  NaN = 1 & Male & 54.45 & 63.56 \\
NaN = 1 & Female &48.52  & 56.33 \\
 NaN = 1 & All &  52.30 & 60.93\\
  ignore Nan & Male & 26.20 & 40.95 \\
 ignore Nan & Female & 23.16 & 34.82 \\
 ignore Nan & All & 25.04 & 38.60 \\

\bottomrule
\end{tabular}
\end{table}

\section{Discussion}
This study tackles the difficulties of segmenting lesions from moderate to severe traumatic brain injuries in T1-weighted MRI scans, where the diversity of lesions complicates conventional methods. By employing a large-scale, multi-dataset supervised pretraining strategy, the proposed model achieved enhanced lesion segmentation performance, exceeding the nnU-Net baseline by as much as 2 Dice points. Unfortunately, the submitted results were adversely affected by a numerical precision error encountered during the ensembling of model predictions. Resolving this issue could have resulted in a performance increase of up to 10 Dice points on the test set. Nonetheless, our further analysis indicated that the model's performance aligns with the dataset's distribution. Specifically, it shows better performance for male patients, who are more frequently represented in the dataset, as well as for age groups that are also more common in the challenge data.

\section*{Acknowledgements}
The authors acknowledge the National Cancer Institute and the Foundation for the National Institutes of Health, and their critical role in the creation of the free publicly available LIDC/IDRI Database used in this study.\\
Part of this work was funded by Helmholtz Imaging (HI), a platform of the Helmholtz Incubator on Information and Data Science.
\clearpage
\bibliographystyle{splncs04}
\bibliography{bibfile.bib}

\clearpage

\end{document}